\documentclass[11pt]{article}

\usepackage[utf8]{inputenc}
\usepackage[T1]{fontenc}
\usepackage[margin=1in]{geometry}
\usepackage{booktabs}
\usepackage{amsmath,amssymb}
\usepackage{graphicx}
\usepackage{xcolor}
\usepackage{pgfplots}
\pgfplotsset{compat=1.18}
\usepackage[numbers,sort&compress]{natbib}
\usepackage[hidelinks]{hyperref}

\title{Compete then Collaborate: Frontier AI Teachers Build a\\
Verifiable Curriculum to Improve a Coding Student Beyond Imitation}
\author{Miseong (Shawn) Kim \\ \small Genesis Cortex AI Inc. \\ \small ORCID: \href{https://orcid.org/0009-0003-7566-3212}{0009-0003-7566-3212}}
\date{July 2026}

\begin{document}
\maketitle

\begin{center}
\small\textbf{Code, data, tests, and verification harness:}
\url{https://github.com/shawnkim678/compete-then-collaborate}\\[-2pt]
\small\emph{Every number is recomputable from the released files or by regenerating teacher solutions under your own provider terms.}
\end{center}

\begin{abstract}
Large language models are increasingly used as \emph{teachers} that generate training
data for smaller \emph{student} models. Prior multi-teacher knowledge distillation merges
several teachers' outputs, but does not ask \textbf{which} frontier model teaches best, and
typically relies on an LLM judge that is known to favor its own outputs. We introduce a
\textbf{compete-then-collaborate} framework in which four frontier AI teachers spanning the major
labs (Claude/Anthropic, Codex-GPT/OpenAI, Grok/xAI, Gemini/Google) are first
\textbf{ranked head-to-head} by an
\emph{execution-based} judge (unit tests / stdin--stdout checks) with fairness controls
(a shared task bank, teacher self-correction, and an intersection-controlled training set),
and then \textbf{collaborate} to build a \emph{verifiable curriculum} for a single coding
student (Qwen2.5-Coder). We report three findings. (1)~Under execution verification, all four
teachers solve standard problems near-perfectly after self-correction ($\approx$99--100\%)
--- a saturation effect, not a skill difference --- while harder competition problems separate
them (Gemini 77\%~$>$~Claude 69\%~$\approx$~Codex 69\%~$>$~Grok 50\%); still, the most robust
results are on the student side and do not depend on the teacher ranking. (2)~\textbf{Imitation
(SFT) on the teachers' verified solutions does not improve --- and can degrade --- an
already-competent coder student} at both 7B and 32B (e.g., 76.7\%$\rightarrow$72.7\% on
MBPP-test, 5.9\%$\rightarrow$2.9\% on competition problems for the union of all teachers).
(3)~The \emph{same} collaborative curriculum used as a \textbf{reinforcement-learning-with-verifiable-rewards
(RLVR)} environment instead \textbf{improves} the student (5.9\%$\rightarrow$8.8\% peak on held-out
competition problems over a 1000-step run, $+49\%$ relative), reversing the direction of SFT. Our central claim is that the value of
AI-teacher collaboration is \textbf{not} pooling answers to imitate, but jointly constructing a
verifiable environment in which the student learns by doing. We release a fully reproducible
on-prem pipeline (NVIDIA GB10) including framework patches required to run GRPO on a
bleeding-edge stack.
\end{abstract}

\section{Introduction}
Distilling capable ``teacher'' LLMs into smaller ``student'' models is now standard practice.
Two questions are under-explored: (i)~\emph{which} commercial frontier model is the better
teacher, measured by the student's real downstream ability rather than by an LLM judge; and
(ii)~whether \emph{combining} teachers is best done by merging their answers or by some other
mechanism.

We study these on Python coding. Our judge is \textbf{code execution} --- objective and free
of the self-preference bias documented for LLM-as-judge setups. We first run a
\textbf{competition}: teachers solve a shared task bank; a teacher's output enters the
student's data only if it passes hidden tests. We then run a \textbf{collaboration}: all
teachers' verified work forms one curriculum, used two ways --- as imitation targets (SFT)
and as a verifiable reward environment (RLVR).

\paragraph{Contributions.}
(1)~An execution-verified, bias-free \textbf{ranking of frontier AI teachers} with three
fairness controls (shared tasks, self-correction, intersection). (2)~A controlled
\textbf{comparison of collaboration modes} showing imitation-SFT fails/degrades competent
coder students while verifiable-reward RL improves them. (3)~A \textbf{reproducible on-prem
(GB10) pipeline}, including the framework patches needed to run GRPO on transformers~5.5 /
torch~2.11 / cu130.

\section{Related Work}
\paragraph{Multi-teacher KD.}
Prior work merges multiple teachers' rationales into a student and reports that naively adding
teachers can \emph{hurt} due to \emph{knowledge conflict}, motivating purification/consolidation
\citep{multiteacherkd}. Our union-SFT result independently reproduces this degradation, but our
framing differs: we do not merge to imitate; we build a verifiable environment.

\paragraph{Teacher choice / LLM-as-judge.}
Studies use Claude and GPT interchangeably as ``strong teachers'' for robustness, and caution
that GPT-4 favors its own generations as an evaluator \citep{rlhfbook}. We remove this bias by
judging with execution.

\paragraph{On-policy distillation \& RLVR.}
On-policy distillation \citep{minillm} and contrastive distillation \citep{distillm2} improve
imitation; RL-with-verifiable-rewards underlies recent reasoning models. We connect these: the
teachers' collaboratively-verified problems become the RLVR reward environment.

\paragraph{Ensemble learning with virtual/synthetic data (historical lineage).}
The intuition that an ensemble of learners trained with \emph{virtual} --- synthetically
generated --- data can generalize beyond any single member predates the deep-learning era:
\citet{jang1999ola} proposed an ensemble learning algorithm driven by virtual data at POSTECH
in 1999. Our setting is a modern realization of the same intuition: the ensemble members are
frontier LLM \emph{teachers}, and the ``virtual data'' is a curriculum of execution-verified
synthetic problems. The connection is one of lineage rather than method --- the 1999 virtual
data augmented ensemble diversity for a single predictor, whereas our verified problems instead
define a \emph{reward landscape} for policy optimization (RLVR). We therefore position this work
as a spiritual successor to that line: the twist is that we do not average ensemble outputs but
use the collaboratively-verified curriculum as a reinforcement-learning environment.

\paragraph{Gap.}
We find no work that ranks \emph{named frontier models as teachers} by execution, contrasts
\emph{imitation vs verifiable-reward collaboration}, and packages it as a
\emph{compete-then-collaborate} pipeline.

\section{Method}
\subsection{Execution-verified generation (the judge)}
Each teacher receives a task (function signature or a competition problem) and returns reasoning
$+$ one code block. We extract the code and run it in an isolated subprocess (rlimits $+$
timeout) against \textbf{hidden tests} (unit-test asserts for function tasks; stdin/stdout for
competition tasks). Only test-passing solutions are kept. Correctness therefore comes from
execution, not from mimicking a teacher.

\subsection{Competition with fairness controls}
(a)~\textbf{Shared task bank} --- all teachers solve identical problems (function-signature
disambiguated; externally-dependent tasks excluded to keep function-only comparison valid).
(b)~\textbf{Self-correction} --- failed tasks are returned to the teacher with the failing test
error for up to two retries, modeling a teacher who revises (raising each teacher's coverage and
removing a ``one-shot luck'' confound). (c)~\textbf{Intersection control} --- the
student-training set is restricted to problems \emph{all} teachers solved, giving identical
problems and equal counts, so only solution/explanation \emph{style} differs.

\subsection{Collaboration: two modes for one student}
The \textbf{union} of all teachers' verified solutions forms the collaborative curriculum. We
use it two ways: \textbf{(SFT)} imitate the pooled solutions; \textbf{(RLVR)} treat the verified
problems as a GRPO reward environment where reward $=$ fraction of tests passed. Student $=$
Qwen2.5-Coder (7B primary, 32B secondary), LoRA.

\section{Experimental Setup}
\begin{itemize}
  \item \textbf{Student}: Qwen2.5-Coder-7B / -32B, LoRA (bf16), GB10 128\,GB unified.
  \item \textbf{Teachers}: Claude, Codex (GPT), Grok, and Gemini via headless CLIs (four major
    labs: Anthropic, OpenAI, xAI, Google). Gemini participates in the execution-verified
    \emph{competition} ranking; the \emph{collaboration} experiments (SFT/RLVR) use the union of
    the first three teachers' verified solutions.
  \item \textbf{Task banks}: function tasks from MBPP (teaching split; MBPP-\emph{test} held
    out), bug-fix tasks via mutation of MBPP references; competition problems from
    \texttt{deepmind/code\_contests} (difficulty 6--9).
  \item \textbf{Held-out eval (no leakage verified)}: MBPP-test (150) and a disjoint competition
    set (68). Metric: execution pass@1.
  \item \textbf{RLVR}: TRL GRPOTrainer $+$ PEFT (HF path; Unsloth kernels avoided due to a
    LoRA-dtype bug); reward $=$ test-pass fraction $+$ small format bonus; HF-generate rollouts
    (vLLM incompatible on this stack).
\end{itemize}

\section{Results}
\subsection{Teacher competition (generation pass@1)}
\textbf{Easy (MBPP, 200 shared problems) --- saturated.}
\begin{table}[h]\centering
\begin{tabular}{lcc}
\toprule
Teacher & one-shot & after self-correction \\
\midrule
Claude & 96.5\% & \textbf{100\%} \\
Grok   & 89.5\% & 99.5\% \\
Codex  & 90.0\% & 99.0\% \\
Gemini & 85.0\% & 99.5\% \\
\bottomrule
\end{tabular}
\caption{Easy MBPP is saturated; all four teachers reach 99--100\% after self-correction, likely
reflecting benchmark exposure rather than skill.}
\end{table}

\textbf{Hard (\texttt{code\_contests}, difficulty 6--9, 150 problems) --- the informative signal.}
\begin{table}[h]\centering
\begin{tabular}{lcc}
\toprule
Teacher & pass@1 (solve) & code-extraction success \\
\midrule
Gemini & \textbf{77\%} (115/150)$^{\ddagger}$ & 100\%$^{\ddagger}$ \\
Claude & 69\% (104/150)\textsuperscript{\S} & 96\% (144/150) \\
Codex  & 69\% (103/150)          & 90\% \\
Grok   & 50\% (75/150)$^{\dagger}$ & 73\% (109/150)$^{\dagger}$ \\
\bottomrule
\end{tabular}
\caption{Hard competition problems separate the teachers: Gemini~$>$~Claude~$\approx$~Codex~$>$~Grok.
Gemini leads by $\approx$8 points; Claude and Codex are within one problem of each other.}
\end{table}

\paragraph{Interpretation (v0.2, addressing peer review).}
MBPP near-ceiling scores (99--100\%) most likely reflect \emph{benchmark saturation / training
exposure} --- MBPP is a public 2021 benchmark seen by all four teachers --- \textbf{not}
differential skill. We therefore do \textbf{not} rest the teacher ranking on MBPP.

On the harder \texttt{code\_contests} set the teachers separate more clearly:
\textbf{Gemini leads (77\%, 115/150), then Claude (69\%)~$\approx$~Codex (69\%), then Grok (50\%)}.
No single vendor's model is uniquely best on hard problems; Gemini is ahead by $\approx$8 points,
while Claude and Codex are within one problem of each other. Crucially, the ranking is \textbf{not}
produced by any LLM judge (execution only), so \emph{self-preference / family bias is excluded by
design}; the residual risks are asymmetric train-leak and CLI/parser artifacts
(Section~\ref{sec:limits}).

\paragraph{Fairness corrections.}
We found and corrected three \emph{infrastructure} artifacts (not capability differences), applied
identically to each affected teacher and reported transparently:
\begin{itemize}
  \item \textbf{\S\ Claude} --- the first run was interrupted at 122/150 problems. We completed the
    remaining 28 (22 passed) so every teacher is scored on the same 150, giving Claude a fair
    \textbf{69\% (104/150)}, up from an incomplete-run 67\% (82/122).
  \item \textbf{\ddag\ Gemini} --- the API's prepaid credits were depleted partway through
    generation, so the final 34/150 problems returned billing errors (429) counted as no-code.
    After the credits were restored we re-ran exactly those 34 and merged them:
    \textbf{77\% (115/150)}, 100\% code-extraction.
  \item \textbf{\dag\ Grok} --- initially 52\% of its batch CLI calls returned empty
    (timeout/flakiness, not wrong code). We added empty-response retries and re-measured:
    \textbf{50\% (75/150)}, extraction 48\%$\to$73\% (109/150). A residual 27\% (41/150) still
    returned no usable code even after retries; we cannot fully separate CLI flakiness from genuine
    difficulty, so we report it transparently rather than discard those items. The remaining
    Grok--Codex/Claude gap (50\% vs 69\%) is thus a conservative (Grok-favorable) estimate.
\end{itemize}

\subsection{Collaboration mode A --- imitation (SFT) does not help competent students}
Held-out pass@1 (intersection-controlled, 197 shared problems):
\begin{table}[h]\centering
\begin{tabular}{lcccc}
\toprule
Student & base & $+$Claude & $+$Codex & $+$Grok \\
\midrule
7B (MBPP)  & 76.7\% & 74.0\% & 70.0\% & 69.3\% \\
32B (MBPP) & 82.0\% & 80.0\% & 77.3\% & 79.3\% \\
\bottomrule
\end{tabular}
\caption{All SFT students fall \emph{below} base; teacher ranking preserved (Claude least
harmful).}
\end{table}

All students fall \textbf{below base}, and the teacher ranking is preserved (Claude least harmful).
The \textbf{union} of all teachers (SFT) is also below base (MBPP 72.7\%, competition \textbf{2.9\%}
vs base 5.9\%).

Imitation at this scale degrades already-competent coder models; the bottleneck is task
\emph{difficulty/headroom}, not model size --- 7B and 32B both degrade.

\subsection{Collaboration mode B --- verifiable-reward RL (RLVR) improves the student}
\label{sec:rlvr}
Same curriculum, competition held-out (base is weak here $\rightarrow$ large headroom):
\begin{table}[h]\centering
\begin{tabular}{lcc}
\toprule
Method & competition base$\rightarrow$student & direction \\
\midrule
SFT (union)              & 5.9\% $\rightarrow$ 2.9\%          & $\downarrow$ degrade \\
RLVR (GRPO, 200 steps)   & 5.9\% $\rightarrow$ 7.4\%          & $\uparrow$ +25\% relative \\
\textbf{RLVR (GRPO, v2, 1000 steps)} & 5.9\% $\rightarrow$ \textbf{8.8\%} peak (7.4\% @1000) & $\uparrow$ \textbf{+49\% rel.\ at peak} \\
\bottomrule
\end{tabular}
\caption{Same data, opposite direction: imitation degrades, verifiable-reward RL improves.
The v2 1000-step run peaks at 8.8\% (step 250--750) with a mild late regression to 7.4\%.}
\end{table}

Training reward rose from $\approx$0 early to 0.25--1.0 late (generalization on train), and
held-out improved. \textbf{Same data, opposite direction from SFT.}

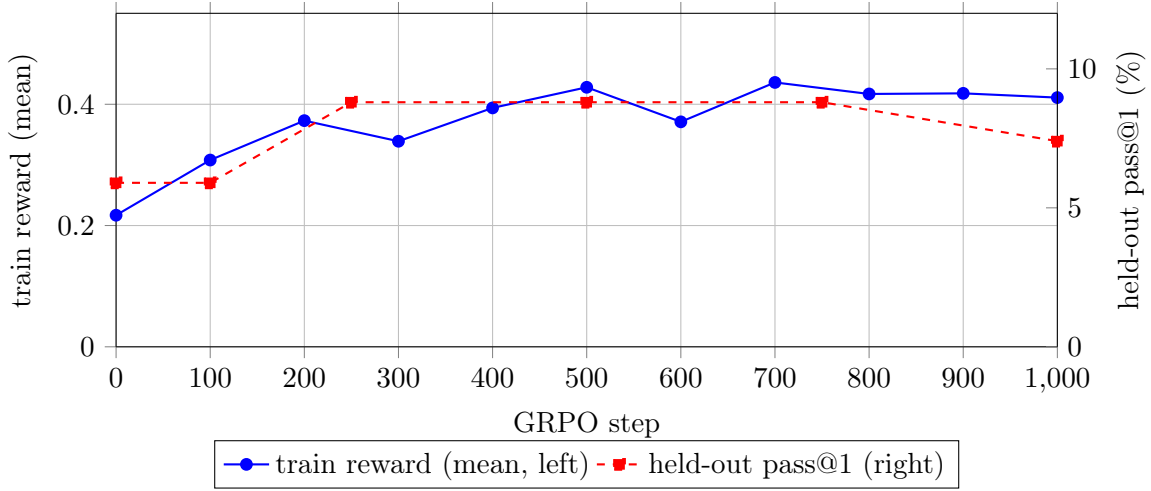
\begin{figure}[h]\centering
\begin{tikzpicture}
\begin{axis}[
    name=ax, width=0.85\linewidth, height=6cm,
    xlabel={GRPO step}, xmin=0, xmax=1000,
    axis y line*=left, ylabel={train reward (mean)}, ymin=0, ymax=0.55,
    grid=both, tick align=outside,
    legend style={at={(0.5,-0.28)}, anchor=north, legend columns=2},
]
\addplot[thick, blue, mark=*] coordinates
  {(0,0.217)(100,0.308)(200,0.373)(300,0.339)(400,0.394)(500,0.428)(600,0.371)(700,0.436)(800,0.417)(900,0.418)(1000,0.411)};
\addlegendentry{train reward (mean, left)}
\addlegendimage{thick, red, dashed, mark=square*}
\addlegendentry{held-out pass@1 (right)}
\end{axis}
\begin{axis}[
    width=0.85\linewidth, height=6cm,
    xmin=0, xmax=1000, hide x axis,
    axis y line*=right, ylabel={held-out pass@1 (\%)}, ymin=0, ymax=12,
]
\addplot[thick, red, dashed, mark=square*] coordinates
  {(0,5.9)(100,5.9)(250,8.8)(500,8.8)(750,8.8)(1000,7.4)};
\end{axis}
\end{tikzpicture}
\caption{v2 learning curve (1000 steps). Left/blue: logged training reward (mean), rising
$0.22\to{\approx}0.43$ then plateauing. Right/red: held-out pass@1 on the 68 competition problems
per checkpoint --- base 5.9\% (4/68), rising to an 8.8\% (6/68) plateau over steps 250--750,
with a mild regression to 7.4\% (5/68) at step 1000. Differences within the plateau correspond to
$\pm1$ problem out of 68 (within sampling noise); the robust signal is base $\to$ RLVR ($4\to5$--$6$/68).}
\end{figure}

\section{Discussion}
The results give a crisp message: \textbf{the value of AI-teacher collaboration is not pooling
answers to imitate, but jointly building a verifiable environment in which the student learns by
doing.} Imitation transfers \emph{style}, which a competent coder already has; RLVR transfers
\emph{capability} by rewarding verified success. This also explains the reproduced
multi-teacher-KD degradation (knowledge conflict): merging answers cannot exceed the student's
imitation ceiling, whereas RL against verifiable rewards can.

\section{Limitations}
\label{sec:limits}
\begin{itemize}
  \item \textbf{Benchmark saturation / train-leak.} MBPP is a widely known 2021 benchmark;
    near-ceiling teacher scores (99--100\%) likely reflect training-set exposure rather than
    differential capability, and our stable-ranking discussion therefore rests on the harder
    \texttt{code\_contests} subset. We do not verify whether individual test items appear
    verbatim in any teacher's pretraining corpus; asymmetric leakage across teachers would bias
    the ranking.
  \item \textbf{Prompt/parser/CLI symmetry.} All teachers share one prompt template and one
    code-extraction parser. We found and corrected a fairness artifact where Grok's batch CLI
    returned empty for 52\% of hard-problem calls (timeout/flakiness, not wrong code); after
    adding empty-response retries the comparison is fair. Residual formatting advantages, if any,
    are bounded by the shared parser and reported extraction-success rates.
  \item \textbf{The teacher ranking is weaker than the student-side result.} On the hard subset
    Gemini leads by $\approx$8 points, while Claude~$\approx$~Codex (within one problem) and Grok
    trails; the fine ordering below Gemini is only weakly separated given the sample size. By
    contrast, the \textbf{SFT-fails / RLVR-succeeds reversal is robust}: it is a change in the
    \emph{student's} held-out performance and does not depend on the teacher ranking or on any LLM judge.
  \item RLVR gains are modest in absolute terms with sparse competition rewards. The 1000-step
    v2 run improves held-out pass@1 from a 5.9\% base to an 8.8\% plateau (steps 250--750) with a
    mild late regression to 7.4\% at step 1000; because the held-out set has only 68 problems,
    differences within the plateau ($\pm1$ problem) are within sampling noise, so we report the
    robust direction (base $\to$ RLVR, $4\to5$--$6$/68) rather than a precise peak, and recommend
    checkpoint selection over training to the final step. A larger held-out set would tighten these
    estimates.
  \item Single student family (Qwen2.5-Coder); one language (Python).
  \item \textbf{Ethics / terms of service.} This is \textbf{academic research} --- a
    benchmarking and methodology study --- not the development, deployment, or distillation of a
    commercial model that competes with any provider; no trained model is offered as a product.
    This purpose places the work outside the \emph{competition-scoped} restrictions of Anthropic
    (no competing product / competing-model training) and OpenAI (no models that compete with
    OpenAI); Google's Gemini terms are likewise competition-scoped. We additionally (i)~do
    \textbf{not} redistribute any teacher's raw outputs, (ii)~do \textbf{not} release models
    trained on teacher outputs, and (iii)~base the headline RLVR result on public competition
    problems with an execution reward --- using \textbf{no teacher-output distillation} --- which
    also addresses xAI's act-scoped restriction on distilling outputs. Correctness is judged by
    deterministic execution, not teacher imitation; released tasks/tests/harness let others
    reproduce every number either offline (re-verify released artifacts) or by regenerating
    teacher solutions under their own provider agreements.
  \item Hardware: a GPU GSP timeout (Xid 119/154) required a reboot during a long RLVR run; we
    mitigate via reduced rollout intensity, checkpointing, and Xid monitoring.
\end{itemize}

\section{Conclusion}
Frontier AIs can be ranked as teachers by an unbiased, execution-based judge, and --- more
importantly --- their collaboration is best expressed as a \emph{verifiable curriculum for
reinforcement learning}, which improves a coding student where imitation fails. We release the
full on-prem pipeline and the framework patches required to reproduce it on NVIDIA GB10.

\section*{Reproducibility}
All code, task banks, hidden tests, results, and this paper are released at
\url{https://github.com/shawnkim678/compete-then-collaborate}. The execution-verified task bank is
also mirrored as a Hugging Face dataset, and the final RLVR (GRPO) LoRA adapter as a Hugging Face
model:
\begin{itemize}
  \item Dataset: \url{https://huggingface.co/datasets/shawnmkim/compete-collab-taskbank}
  \item Model:   \url{https://huggingface.co/shawnmkim/qwen2.5-coder-7b-rlvr-compete-collab}
\end{itemize}
Readers can re-verify every reported number offline
(\texttt{python reproduce.py --check-banks --selftest} on the released artifacts) or regenerate
teacher solutions with their own API keys under their own provider terms; we redistribute no
teacher outputs (see the ethics statement).

\medskip\noindent
Scripts: \texttt{verify\_code.py},
\texttt{verify\_stdio.py}, \texttt{build\_taskbank\_mbpp.py}, \texttt{build\_contests\_bank.py},
\texttt{prof\_run.py}, \texttt{prof\_run\_stdio.py}, \texttt{prof\_retry.py},
\texttt{equalize\_golden.py}, \texttt{eval\_code\_students.py}, \texttt{grpo\_train.py}.
Orchestration (setsid): \texttt{grpo\_orchestrator.sh}. GB10 framework patches (trl import
flags, vllm-off, \texttt{PreTrainedModel.warnings\_issued}, HF$+$PEFT to avoid the Unsloth
LoRA-dtype bug) documented in the research log \S16--17.

\section*{Acknowledgments}
We thank Dr.\ Min Jang (Pohang University of Science and Technology, POSTECH),
whose 1999 dissertation on ensemble learning with virtual data \citep{jang1999ola}
provided the conceptual lineage for the verifiable-curriculum framing developed here.
In his review of an earlier draft, Dr.\ Jang highlighted the engineering contributions
of the reported system as its principal strengths --- the stack-level debugging on the
NVIDIA GB10 platform, the CLI orchestration across four commercial LLMs, and the
execution sandbox. No third-party copyrighted material is redistributed with this
paper; the referenced dissertation is cited bibliographically only.

\bibliographystyle{plainnat}
\bibliography{refs}

\end{document}